\documentclass[10pt,twocolumn,letterpaper]{article}

\usepackage[dvipsnames,table]{xcolor}

\usepackage[pagenumbers]{iccv} 

\usepackage[accsupp]{axessibility}
\usepackage{bm}
\usepackage{multirow}
\usepackage{algorithm,algorithmic}
\usepackage{amssymb}

\usepackage{listings}
\definecolor{codegreen}{rgb}{0,0.6,0}
\definecolor{codegray}{rgb}{0.5,0.5,0.5}
\definecolor{codepurple}{rgb}{0.58,0,0.82}

\lstdefinestyle{mystyle}{
    backgroundcolor=\color{white},
    commentstyle=\color{codegreen},
    keywordstyle=\color{magenta},
    numberstyle=\tiny\color{codegray},
    stringstyle=\color{codepurple},
    basicstyle=\ttfamily\footnotesize,
    breakatwhitespace=false,
    breaklines=true,
    captionpos=b,
    keepspaces=true,
    numbersep=5pt,
    showspaces=false,
    showstringspaces=false,
    showtabs=false,
    tabsize=2
}
\definecolor{iccvblue}{rgb}{0.21,0.49,0.74}
\usepackage[pagebackref,breaklinks,colorlinks,allcolors=iccvblue]{hyperref}

\def\paperID{5231} 
\def\confName{ICCV}
\def\confYear{2025}

\title{Dynamic-DINO: Fine-Grained Mixture of Experts Tuning for Real-time Open-Vocabulary Object Detection}

\newcommand*\samethanks[1][\value{footnote}]{\footnotemark[#1]}

\author{
Yehao Lu$^1$\thanks{The first two authors contributed equally to this paper.}, 
Minghe Weng$^1$\samethanks[1],
Zekang Xiao$^2$\samethanks[1],
Rui Jiang$^1$,
Wei Su$^1$,
Guangcong Zheng$^1$,\\
Ping Lu$^3$,
Xi Li$^{1,2}$\thanks{Corresponding author.}\\
$^1$College of Computer Science and Technology, Zhejiang University \\
$^2$Polytechnic Institute, Zhejiang University \enspace $^3$ZTE \\
{\tt\small \{luyehao, wengminghe, xiaozekang, jrss, weisuzju, guangcongzheng, xilizju\}@zju.edu.cn}\\
{\tt\small Lu.ping@zte.com.cn}
}

\begin{document}

\twocolumn[{%
\renewcommand\twocolumn[1][]{#1}%
\maketitle
\vspace{-9mm}

\begin{center}
    \centering
    \captionsetup{type=figure}
    \includegraphics[width=0.98\textwidth]{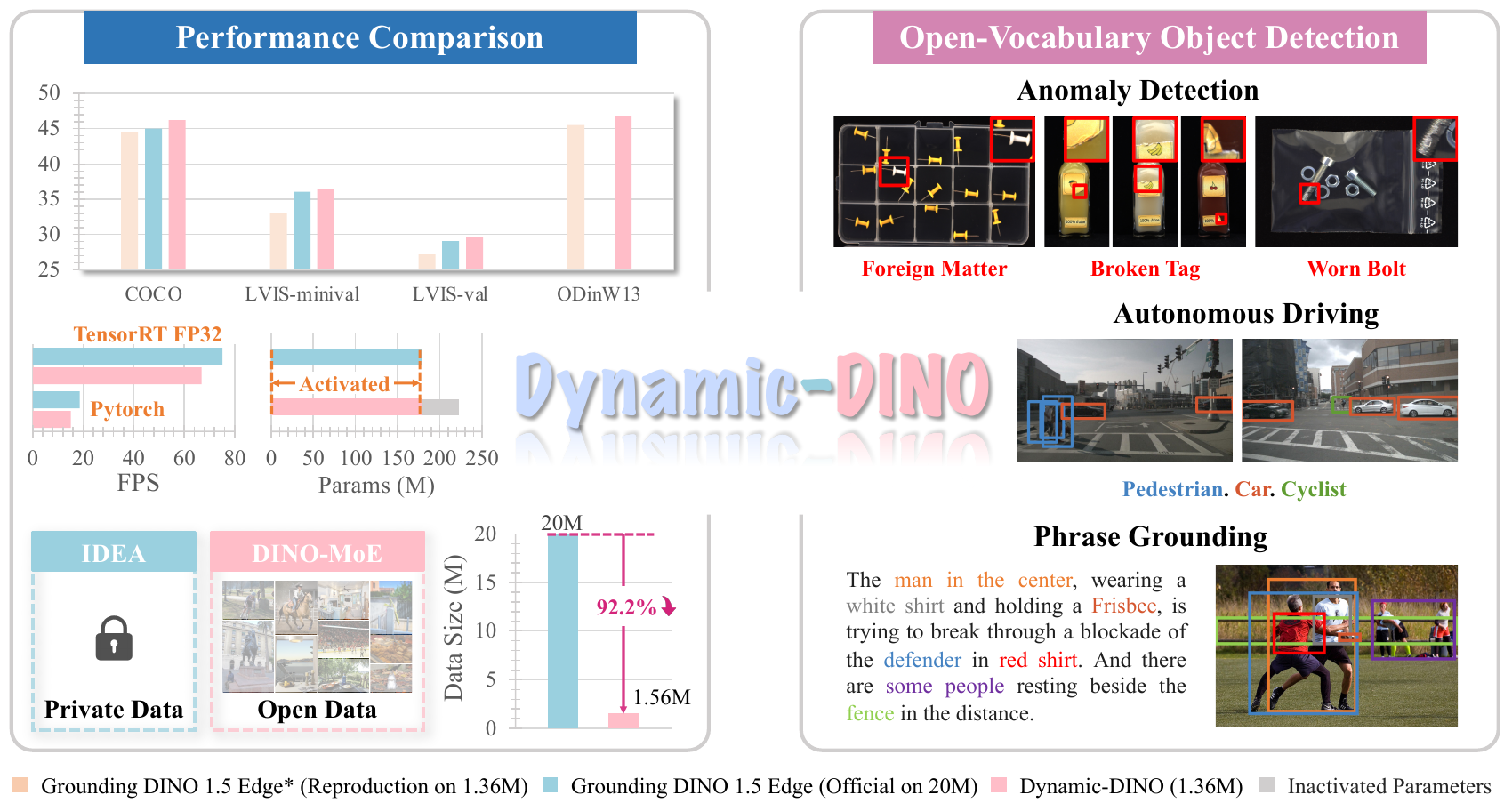}
    \caption{Dynamic-DINO is an efficient object-centric vision model designed for open-vocabulary object detection. Pretrained with merely 1.56M open-source data, Dynamic-DINO outperforms Grounding DINO 1.5 Edge, which is pretrained on the private Grounding20M dataset, across multiple zero-shot benchmarks. Furthermore, we have rigorously constrained the number of activated parameters during inference to align with that of Grounding DINO 1.5 Edge, ensuring comparable inference speed.}
    \label{first_img}
\end{center}
}]

\def\thefootnote{*}\footnotetext{Equal contribution.}
\def\thefootnote{$\dagger$}\footnotetext{Corresponding author.}

\maketitle
\begin{abstract}
The Mixture of Experts (MoE) architecture has excelled in Large Vision-Language Models (LVLMs), yet its potential in real-time open-vocabulary object detectors, which also leverage large-scale vision-language datasets but smaller models, remains unexplored. This work investigates this domain, revealing intriguing insights. 
In the shallow layers, experts tend to cooperate with diverse peers to expand the search space. While in the deeper layers, fixed collaborative structures emerge, where each expert maintains 2-3 fixed partners and distinct expert combinations are specialized in processing specific patterns.
Concretely, we propose Dynamic-DINO, which extends Grounding DINO 1.5 Edge from a dense model to a dynamic inference framework via an efficient MoE-Tuning strategy. 
Additionally, we design a granularity decomposition mechanism to decompose the Feed-Forward Network (FFN) of base model into multiple smaller expert networks, expanding the subnet search space.
To prevent performance degradation at the start of fine-tuning, we further propose a pre-trained weight allocation strategy for the experts, coupled with a specific router initialization.
During inference, only the input-relevant experts are activated to form a compact subnet. Experiments show that, pretrained with merely 1.56M open-source data, Dynamic-DINO outperforms Grounding DINO 1.5 Edge, pretrained on the private Grounding20M dataset. The code will be publicly available at \href{https://github.com/wengminghe/Dynamic-DINO}{https://github.com/wengminghe/Dynamic-DINO}.
\end{abstract}    
\begin{figure*}[!t] 
    \centering
    \includegraphics[width=1.0\textwidth]{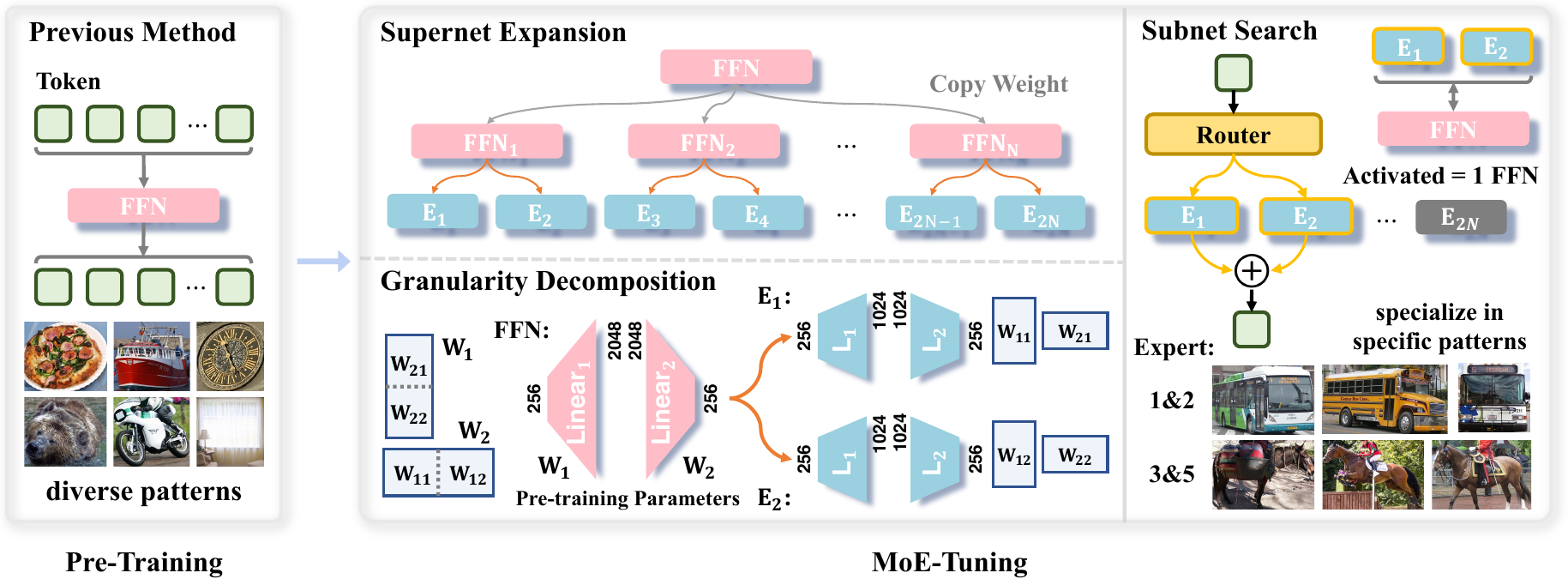}
    \vspace{-0.6cm}
    \caption{\textbf{Illustration of Dynamic-DINO.} In previous transformer blocks, a single FFN handles diverse token patterns, causing gradient conflicts and long-tail issues. MoE-Tuning extends the dense model into a sparse dynamic inference framework, activating only relevant experts to form a compact subnet during inference. Experiments show that deeper layers develop stable expert collaboration, with specialized combinations for specific token patterns. Finer expert granularity enhances specialization, prompting the introduction of granularity decomposition for fine-grained expert segmentation. To align with MoE-Tuning, we further propose a pre-trained weight allocation strategy for the experts to prevent performance degradation at the start of fine-tuning.}
    \label{Intro}
    \vspace{-0.3cm}
\end{figure*}

\vspace{-0.3cm}

\section{Introduction}
\label{sec:intro}

In recent years, open-vocabulary object detection \cite{GLIP,Detclip,minderer2023scaling,Zhang_2023_ICCV,wang2023detecting,T-rex2} has emerged as a pivotal paradigm for foundational vision tasks. 
In contrast to general object detectors \cite{ren2015faster} which are limited to detecting objects within predefined and fixed categories, such models flexibly localize arbitrary objects with the integration of language modality. 
Notably, real-time open-vocabulary object detectors \cite{Yolo-world,YOLO-UniOW,GroundingDINO1.5,DINOX} have garnered increasing emphasis due to their significant practical value, having been widely applied in various fields \cite{lu2023scene, han2021feedback}, such as anomaly detection, robotics and autonomous driving.

Current real-time open-vocabulary object detectors \cite{Yolo-world,YOLO-UniOW,GroundingDINO1.5,zhao2024real,wang2024ovlw} mainly adopt dense models with fixed inference architectures. 
In contrast, Mixture of Experts (MoE) \cite{lepikhin2020gshard,lin2024moe,dai2024deepseekmoe,bai2023qwen} activates only a subset of the neural network during inference to simultaneously scale up model capacity and ensure efficient computation, which is highly compatible with this field, yet their integration remains under-explored. 
From another perspective, MoE has demonstrated success in Large Vision-Language Models (LVLMs) \cite{lin2024moe,STGC}. Similarly, real-time open-vocabulary object detectors are trained on large-scale vision-language datasets but with reduced model scales. Exploring the potential of MoE in such compact multimodal models is an intriguing issue as well.
Thus, this work investigates this domain.

Concretely, MoE replaces the feed-forward network (FFN) in each transformer layer with multiple expert networks, scaling up model capacity to enhance performance. During inference, it employs a router to activate only a subset of experts, ensuring efficient computation.
In previous object detectors, a single FFN in each layer is required to process all tokens, which encompass extensive patterns in open scenarios, including visual patterns (e.g., category and attribute) and contextual patterns (e.g., relative position and relationship). This not only slows down model learning but also leads to gradient conflicts and long-tail issues. When exploring the MoE approach, we observe that deeper layers develop stable expert collaboration, with specialized combinations for specific token patterns, as illustrated in Fig. \ref{Intro}.
Intuitively, finer expert granularity expands the subnet search space, enabling MoE to partition input tokens more precisely. This simplifies model learning, allowing a powerful network to be trained with relatively limited data. Thus, efficiently expanding the search space is crucial.

For a MoE network with $N$ experts, where the top-$k$ experts are activated during inference, the search space size is $(C_N^k)^L$, where $L$ represents the number of layers.
To expand the search space, there are intuitively two ways. 
First, increasing the number of activated experts $k$. However, this approach inevitably leads to higher computational costs during inference. 
Second, increasing the number of experts $N$. Yet, this approach results in higher memory costs and slower training speeds. Additionally, when the amount of training data is limited, it may cause overfitting issues. 

To address this challenge, we propose a novel dynamic inference framework, namely Dynamic-DINO, for real-time open-vocabulary object detection. 
For cost efficiency, we adopt an efficient fine-tuning paradigm based on the reproduced Grounding DINO 1.5 Edge.
Following MoE, we replicate the FFN in the Transformer layer $N$ times to expand model parameters, forming a supernet, while initializing the extended FFNs with pretrained FFN parameters.
Inspired by DeepSeekMoE \cite{dai2024deepseekmoe}, we introduce a granularity decomposition strategy, which splits a single FFN into multiple expert networks. Distinctly, we decompose the FFN's parameters and allocate them to initialize the expert networks, ensuring the sum of expert network outputs matches the FFN output for each token.
This approach increases the number of experts without enlarging the total parameter count, effectively expanding the search space.
During feed-forward inference, a router network is utilized to selectively activate a subset of experts, forming a compact subnet, while strictly maintaining activated parameters equivalent to a single FFN.

To validate the effectiveness of our method, we evaluate its zero-shot performance on multiple benchmarks, including COCO \cite{coco}, LVIS \cite{LVIS} and ODinW \cite{GLIP}.
Training with merely 1.56M open-source data comprising Object365 \cite{O365}, GoldG \cite{GoldG} and V3Det \cite{V3Det} datasets, Dynamic-DINO outperforms Grounding DINO 1.5 Edge, which is pretrained on the private Grounding20M dataset, with comparable inference speed. 
To facilitate further research, we emphasize reproducibility and accessibility.

Our contributions can be summarized as:
\begin{itemize}
    \item We validate the potential of integrating the MoE into the real-time open-vocabulary object detection task.
    \item We propose a novel MoE-Tuning method that, through granularity decomposition of the FFN, expands the search space while keeping the parameter count constant, facilitating effective modeling of the extensive patterns.
    \item Our method surpasses Grounding DINO 1.5 Edge with merely 1.56M open-source training data with comparable inference speed. 
\end{itemize}

\section{Related Work}
\label{sec:RelatedWork}

\subsection{Open-Vocabulary Object Detection}
\vspace{-0.1cm}
Open-vocabulary object detection \cite{Zareian_2021_CVPR, Gu_2022_ICLR} has consistently attracted the community’s attention. Representative works include GLIP \cite{GLIP}, OpenSeeD \cite{Zhang_2023_ICCV}, OWL-ViT \cite{minderer2022simple}, OWL-ST \cite{minderer2023scaling}, Grounding DINO \cite{GroundingDINO1.0}, DetCLIP \cite{yao2022detclip, Yao_2023_CVPR, Yao_2024_CVPR}, OV-DINO \cite{OV-DINO}, UniDetector \cite{wang2023detecting}, to name a few. 
Notably, real-time detectors have garnered increasing emphasis. 
YOLO-World \cite{Yolo-world} and YOLO-UniOW \cite{YOLO-UniOW} inherit the efficient computational capabilities of the YOLO series \cite{yolo,YOLO9000,Yolov3} detectors and extend them to the open-vocabulary domain.
Grounding DINO 1.5 \cite{GroundingDINO1.5} proposes the Edge model, focusing on computational efficiency.
Grounding DINO 1.6 and DINO-X \cite{DINOX} further enhance performance by expanding the pre-training dataset based on the Grounding DINO 1.5 Edge.
Additionally, OmDet-Turbo \cite{zhao2024real} and OVLW-DETR \cite{wang2024ovlw} have also achieved real-time detection.
Distinct from the aforementioned methods, we innovatively incorporate MoE-driven dynamic inference to achieve significant improvements in accuracy without compromising efficiency.




\vspace{-0.2cm}
\subsection{Mixture of Experts}
\vspace{-0.1cm}
Mixture-of-Experts (MoE) is a prominent architecture in conditional computation \cite{han2021dynamic,wang2022sp,wang2018skipnet,veit2018convolutional}, which has shown potential in scaling up models \cite{shazeer2017outrageously}. The core principle of MoE lies in the use of a router that allocates tokens to experts. 
Early works have adopted the hard routing mode \cite{bao2022vlmo,wang2023image,shen2023scaling,liang2022mind}, where each expert is typically assigned a specific role. In contrast, recent LLM and LVLM works have focused on soft routers, which enables a dynamic allocation of tokens among different experts, including Gshard \cite{lepikhin2020gshard}, Lifelong-MoE \cite{chen2023lifelong}, MoE-LLaVA \cite{lin2024moe}, LLaVA-MoLE \cite{chen2024llava}, MoCLE \cite{gou2023mixture}, DEMIX \cite{gururangan2021demix}, to name a few. 
Among these, DeepSeekMoE \cite{dai2024deepseekmoe} and QwenMoE \cite{bai2023qwen} segment experts by splitting the FFN intermediate hidden dimension. We adopt this latest design, but with a key distinction. Unlike their approach of randomly initializing experts for full pre-training, we generate experts by segmenting pre-trained FFN parameters for incremental fine-tuning. Another key contribution of our work is validating the effectiveness of MoE fine-tuning in open-vocabulary object detection.
\begin{figure*}[!t] 
    \centering
    \includegraphics[width=0.98\textwidth]{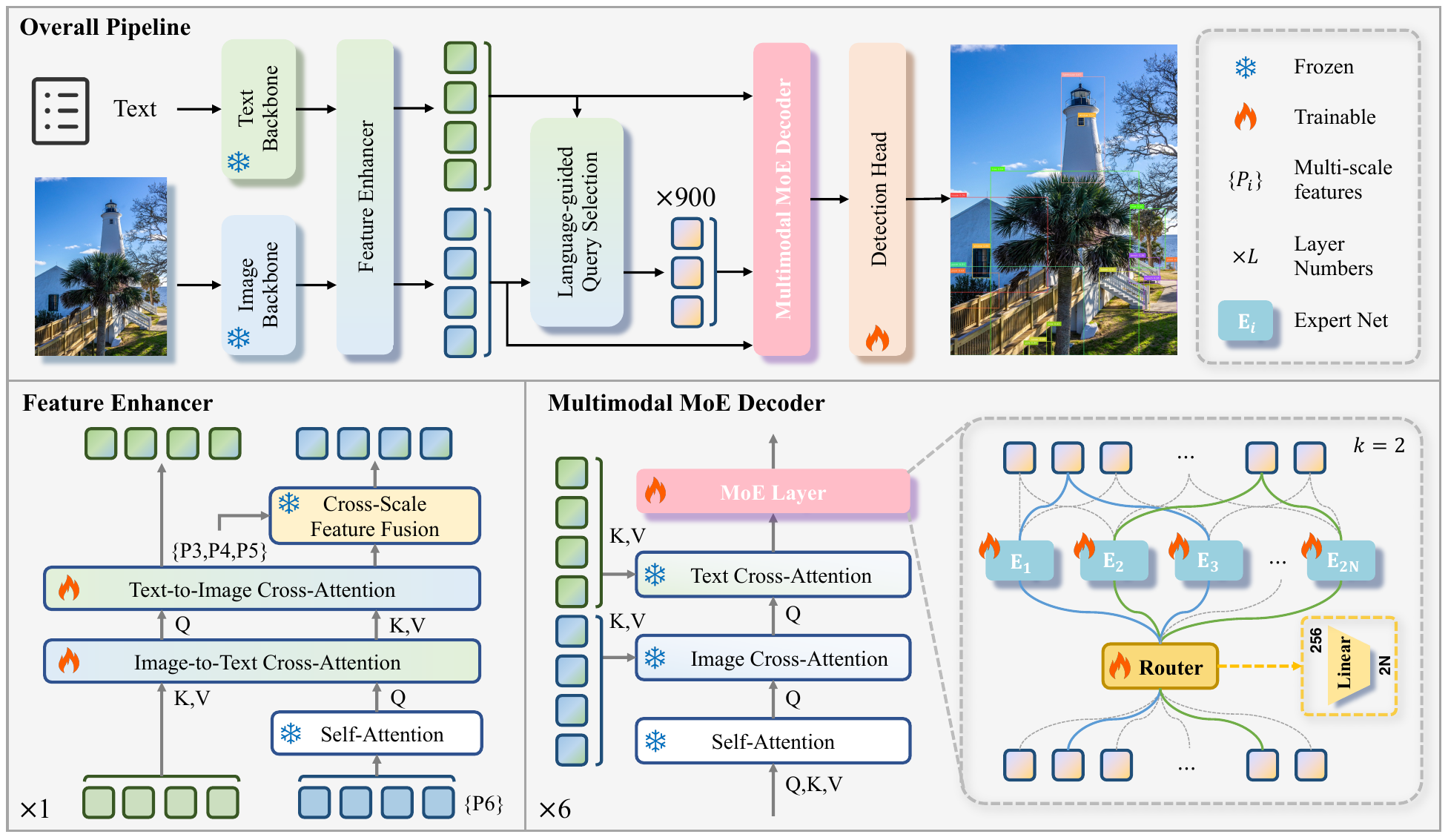}
    \caption{\textbf{MoE-Tuning framework.} Dynamic-DINO builds upon the Grounding DINO 1.5 Edge \cite{GroundingDINO1.5}, extending it from a dense model into a dynamic inference framework via the proposed MoE-Tuning strategy.}
    \label{Method}
    \vspace{-0.3cm}
\end{figure*}

\vspace{-0.2cm}
\section{Methods}
\label{sec:Methods}


\subsection{Overview}
\label{Overview}
\vspace{-0.1cm}
The overall pipeline is depicted in Fig. \ref{Method}. 
Dynamic-DINO builds upon the Grounding DINO 1.5 Edge \cite{GroundingDINO1.5}, extending it from a dense model into a dynamic inference framework via MoE-Tuning.
Due to its closed-source status, we have reproduced and trained the base model on publicly available datasets. 
For MoE-Tuning, we employ the sparse MoE structure to the decoder, for two reasons. First, after the Language-guided Query Selection, only 900 tokens are retained, significantly fewer than in previous modules, which minimizes the computational costs introduced by the router selection. Second, the final output of the decoder directly influences bounding box regression, making it more efficient for fine-tuning.
To balance accuracy and training efficiency during MoE-Tuning, we allow the Cross-Attention in the Feature Enhancer, the MoE Layer in the Cross-Modality MoE Decoder, and the Detection Head to participate in training, while freezing all other parameters.

\subsection{Cross-Modality MoE Decoder}
\noindent \textbf{Supernet Expansion.} 
Following MoE \cite{fedus2022switch} paradigm, we scale up the model by expanding the FFN in each layer of the decoder into $N$ FFNs of identical size.
For each FFN, its intermediate hidden dimension is evenly divided into $k$ partitions, thereby constructing $k\times N$ experts. Fig. \ref{Intro} presents the case where $k=2$.
In this manner, the model's capacity is expanded to form a supernet. Meanwhile, the finer granularity of experts leads to a larger search space for subnets.

\noindent \textbf{Subnet Inference.}
During feed-forward inference, the router $\mathrm{R}(x)$ serves as the critical component for subnet selection, which is a single linear layer as shown in Fig. \ref{Method}, where $x$ is the input token. 
Its output is normalized by the softmax function to obtain the score $s = [s_1,s_2,...,s_{kN}] \in \mathbb{R}^{kN}$ for each expert, which can be formulated as:
\begin{equation}
    s_i = \frac{e^{R(x)_i}}{\sum_{j=1}^{kN} e^{R(x)_j}}
\end{equation}
Next, the top-$k$ experts with the highest scores are selected for activation through a gating mechanism, ensuring that the activated parameters remain equivalent to those of a single FFN. The gate $g \in \mathbb{R}^{kN}$ is calculated as:
\begin{equation}
    g_i=\left\{
    \begin{array}{l}
        1, \quad s_i \in \mathrm{Topk}(\{s_j|0\le j<kN\},k), \\
        0, \quad \mathrm{otherwise},
    \end{array}
    \right.
    \label{gi}
\end{equation}
The output of the Sparse MoE Layer $h(x)$ is the sum of the outputs from the selected experts $\mathrm{E}_i$, which satisfies $g_i=1$. For formal clarity, this process is expressed as:
\begin{equation}
    h(x)=\sum_{i=1}^{kN}g_i \cdot \mathrm{E}_i(x)
    \label{ffn}
\end{equation}




\subsection{MoE-Tuning}
\label{MoE-Tuning}
\noindent \textbf{Expert Initialization.} Each FFN is initialized with the parameters from the pre-trained base model, which consists of two linear layers, denoted as $[W_1,b_1,W_2,b_2]$, where $W_1 \in \mathbb{R}^{H\times D}$, $b_1 \in \mathbb{R}^{H\times 1}$, $W_2 \in \mathbb{R}^{D\times H}$, $b_2 \in \mathbb{R}^{D\times 1}$, $D$ denotes the input token dimension, and $H$ represents the hidden layer dimension of the FFN.
The feed-forward process of FFN is calculated as:
\begin{equation}
    \mathrm{FFN}(x)= W_2(\sigma (W_1x+b_1))+b_2
    \label{ffn}
\end{equation}
where $x \in \mathbb{R}^{D\times 1}$ and $\sigma$ is activation function. The parameters of each fine-grained expert are further segmented based on each FFN. Specifically, the parameters of the first linear layer is horizontally divided into $k$ blocks as follows:
\begin{equation}
    W_1=\{W_1^i \in \mathbb{R}^{(H/k)\times D} | 1\le i \le k\}
\end{equation}
\vspace{-0.3cm}
\begin{equation}
    b_1=\{b_1^i \in \mathbb{R}^{(H/k)\times 1} | 1\le i \le k\}
\end{equation}
\begin{figure}[!t] 
    \centering
    \includegraphics[width=0.48\textwidth]{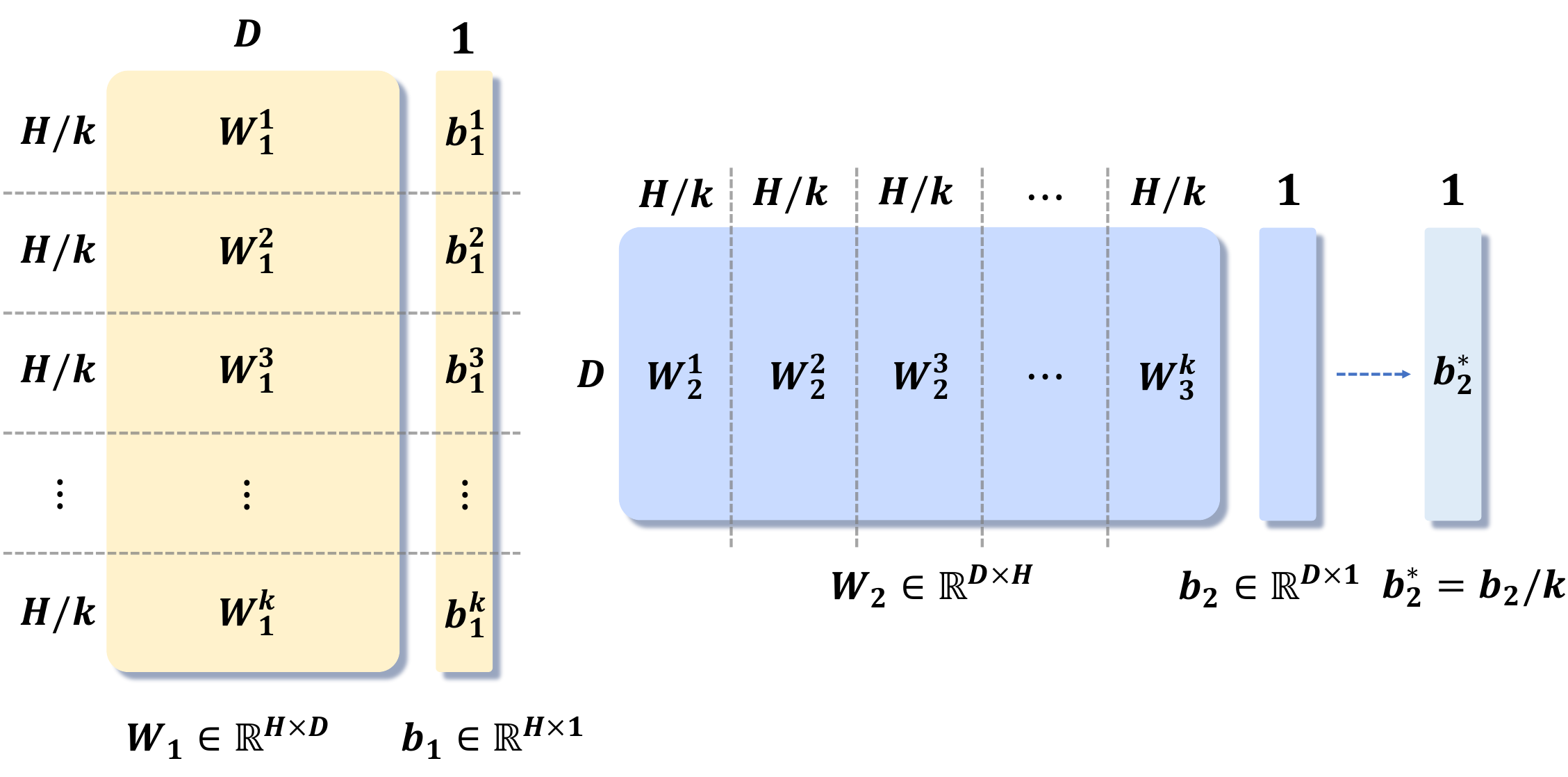}
    \caption{\textbf{Expert initialization.} We decompose the parameters of pre-trained FFN and allocate them to initialize the multiple expert networks, ensuring that the sum of the outputs from the $k$ fine-grained experts matches the output of the pre-trained FFN.}
    \label{split}
\end{figure}
Next, the parameters of the second linear layer is vertically divided as:
\begin{equation}
    W_2=\{W_2^i \in \mathbb{R}^{D\times (H/k)} | 1\le i \le k\}
\end{equation}
\vspace{-0.3cm}
\begin{equation}
    b_2^*=b_2/k
\end{equation}
The $i$-th expert $\mathrm{E}_i$ is formally a smaller FFN, with parameters $[W_1^i, b_1^i, W_2^i, b^*]$. 
This weight allocation strategy is illustrated in Fig. \ref{split}.
This parameter segmentation ensures that the sum of the outputs from the $k$ fine-grained experts matches the output of the original FFN:
\begin{equation}
    \mathrm{FFN}(x)=\sum_{j=1}^{k}\mathrm{E}_j(x)
\end{equation}

\begin{figure}[!t] 
    \centering
    \includegraphics[width=0.48\textwidth]{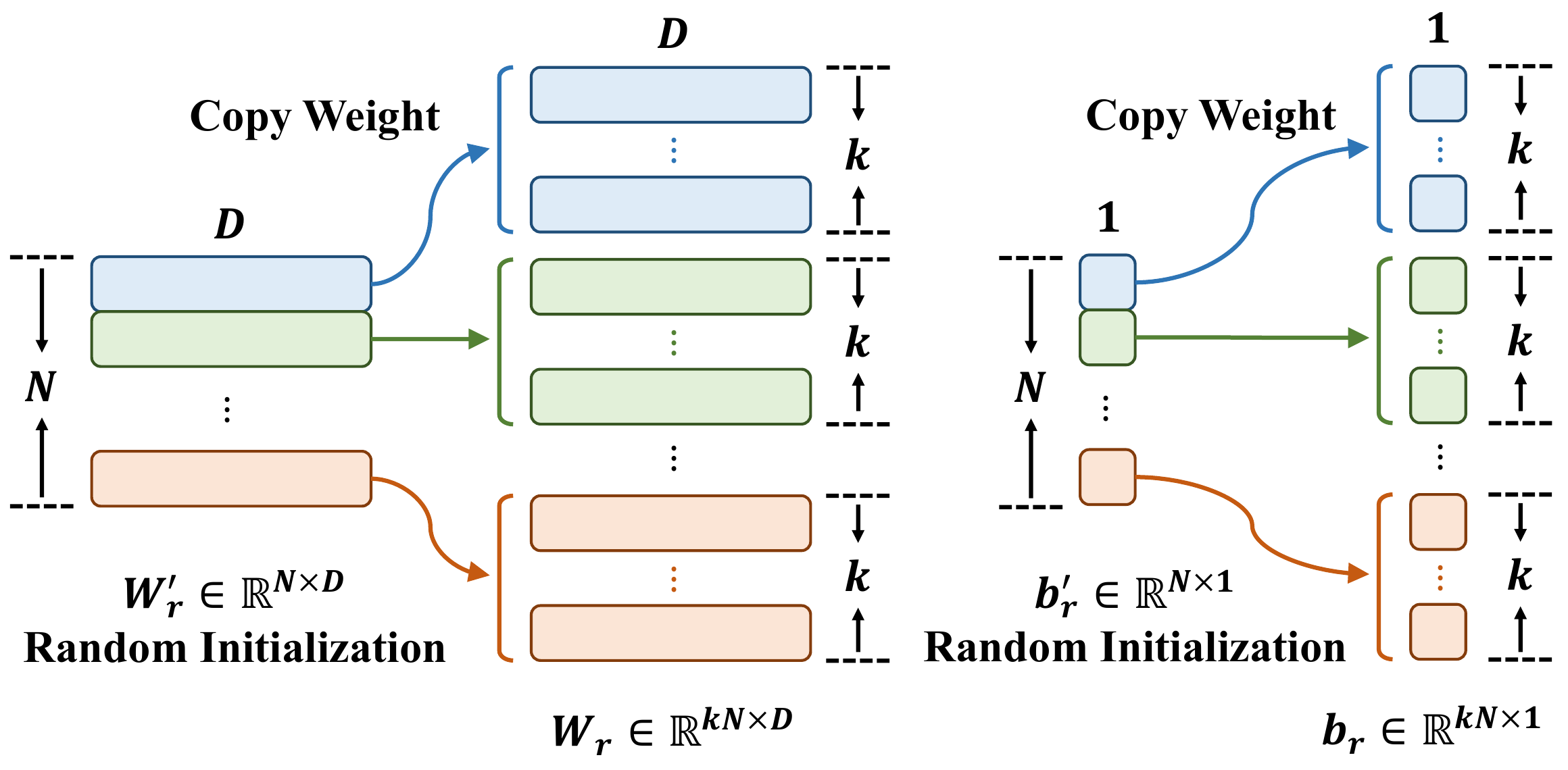}
    \caption{\textbf{Router initialization.} This initialization ensures that, at the beginning of fine-tuning, the router invariably selects the $k$ experts derived from the same FFN, enabling incremental performance improvements over the base model, preventing abrupt performance degradation.}
    \label{router}
\end{figure}

\begin{figure}[!t] 
    \centering
    \includegraphics[width=0.48\textwidth]{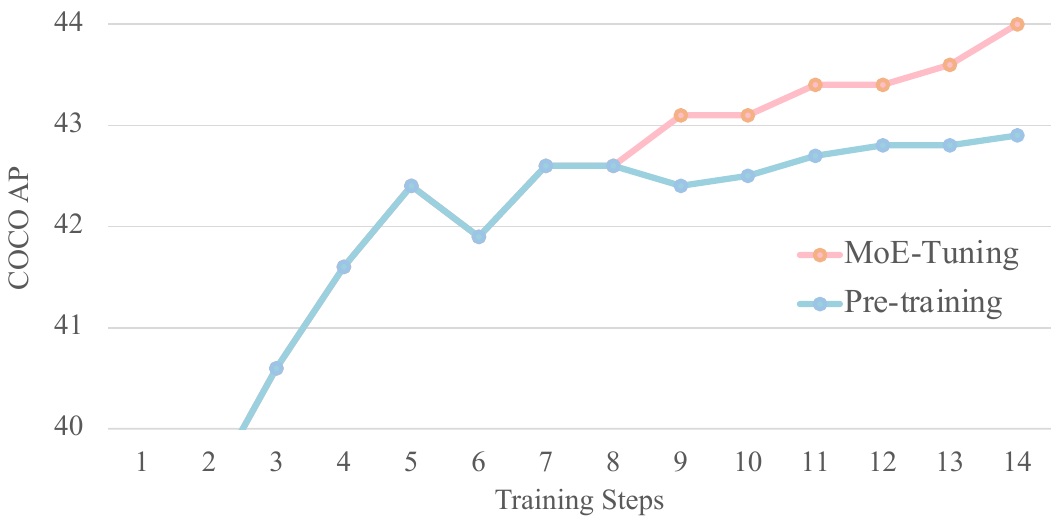}
    \caption{\textbf{Effect of MoE-Tuning.} Based on specially designed expert and router initialization methods, MoE-Tuning ensures incremental performance improvement. The results on COCO with 640 × 640 resolution demonstrate that MoE-Tuning provides significant performance enhancements compared to pre-training.}
    \label{training_coco}
\end{figure}

\noindent \textbf{Router Initialization.} 
The router is implemented as a single linear layer, with its parameters denoted as $[W_r, b_r]$, where $W_r \in \mathbb{R}^{kN\times D}$ and $b_r \in \mathbb{R}^{kN\times 1}$. 
To achieve incremental performance improvement on the base model during fine-tuning, it is essential to ensure that the sum of the outputs from the initial activated experts precisely match the output of the pre-trained FFN, i.e., $h(x)=\mathrm{FFN}(x)$. 
Consequently, specific constraints must be imposed on the router initialization.
As shown in Fig. \ref{router}, we first randomly initialize the weights $W'_r \in \mathbb{R}^{N\times D}$ and $b'_r \in \mathbb{R}^{N\times 1}$, and then replicate each centroid vector in $W'_r$ and $b'_r$ $k$ times to form the router weights $W_r$ and $b_r$.
With this initialization, the router is guaranteed to select the $k$ experts derived from the same FFN at the start of fine-tuning.
As shown in Fig. \ref{training_coco}, our method achieves incremental performance improvements during fine-tuning.




\begin{table*}[!t]
  \centering
  \caption{\textbf{Comparison of zero-shot performance on COCO, LVIS-minival, and LVIS-val object detection benchmarks.} Dynamic-DINO×16-Top2 model is utilized, which comprises $kN=16$ experts and activates $k=2$ experts. Grounding DINO 1.5 Edge* indicates the results of our replication, which also serves as our base model.}
  \resizebox{0.99 \textwidth}{!}{
    \begin{tabular}{lllccccccccccc}
    \toprule
    \multicolumn{1}{l}{\multirow{2}[2]{*}{Method}} & \multicolumn{1}{l}{\multirow{2}[2]{*}{Backbone}} & \multicolumn{1}{l}{\multirow{2}[2]{*}{Pre-training Data}} & \multirow{2}[2]{*}{Test Size} & COCO-val & \multicolumn{4}{c}{LVIS-minival} &       & \multicolumn{4}{c}{LVIS-val} \\
\cmidrule{6-9}\cmidrule{11-14}          &       &       &       & $\rm AP_{box}$ &  $\rm AP_{all}$ &$\rm AP_{r}$ & $\rm AP_{c}$ & $\rm AP_{f}$ &       & $\rm AP_{all}$ & $\rm AP_{r}$ & $AP_{\rm c}$ & $AP_{\rm f}$ \\
    \hline
    \rowcolor{gray!15} \multicolumn{14}{l}{End-to-End Open-Set Object Detection} \\
    GLIP \cite{GLIP}  & Swin-T & O365,GoldG,Cap4M & 800 × 1333 & 46.3  & 26.0    & 20.8  & 21.4  & 31.0    &       & -     & -     & -     & -  \\
    Grounding DINO \cite{GroundingDINO1.0} & Swin-T & O365,GoldG,Cap4M & 800 × 1333 & 48.4  & 27.4  & 18.1  & 23.3  & 32.7  &       & -     & -     & -     & -   \\
    \hline
    \rowcolor{gray!15} \multicolumn{14}{l}{Real-time End-to-End Open-Set Object Detection Models} \\
    YOLO-Worldv2-S \cite{Yolo-world} & YOLOv8-S & O365,GoldG & 640 × 640 & -     & 22.7  & 16.3  & 20.8  & 25.5  &       & 17.3  & 11.3  & 14.9  & 22.7 \\
    YOLO-Worldv2-M \cite{Yolo-world} & YOLOv8-M & O365,GoldG & 640 × 640 & -     & 30.0    & 25.0    & 27.2  & 33.4  &       & 23.5  & 17.1  & 20.0    & 32.6  \\
    YOLO-Worldv2-L \cite{Yolo-world} & YOLOv8-L & O365,GoldG & 640 × 640 & -     & 33.0    & 22.6  & 32    & 35.8  &       & 26.0    & 18.6  & 23    & 32.6  \\
    YOLO-Worldv2-L \cite{Yolo-world} & YOLOv8-L & O365,GoldG,CC3M-Lite & 640 × 640 & -     & 32.9  & 25.3  & 31.1  & 35.8  &       & 26.1  & 20.6  & 22.6  & 32.3  \\
    OmDet-Turbo-T \cite{zhao2024real} & Swin-T & O365,GoldG & 640 × 640 & 42.5  & 30.0    & -     & -     & -     &       & -     & -     & -     & -      \\
    OVLW-DETR-L \cite{wang2024ovlw} & LW-DETR-L & O365,GoldG & 640 × 640 & -     & 33.5  & 26.5  & 33.9  & 34.3  &       & -     & -     & -     & -    \\
    \hline
    \rowcolor{gray!15} \multicolumn{14}{l}{Efficient Object-Centric Vision Model} \\
    Grounding DINO 1.5 Edge \cite{GroundingDINO1.5} & EfficientViT-L1 & Grounding-20M & 640 × 640 & 42.9  & 33.5  & 28.0    & 34.3  & 33.9  &       & 27.3  & 26.3  & 25.7  & 29.6  \\
    Grounding DINO 1.5 Edge* & EfficientViT-L1 & O365,GoldG,V3Det ($\rm \approx1.56 M$) & 640 × 640 & 42.6  & 31.1  & 33.8  & 34.3  & 27.8  &       & 25.4  & 31.8  & 24.8  & 23.3   \\
    Dynamic-DINO (Ours) & EfficientViT-L1 & O365,GoldG,V3Det ($\rm \approx1.56 M$) & 640 × 640 &  43.7  &  33.6 & 37.0  &  36.6  & 30.3 &       &  27.4   &  32.4 & 26.9 & 25.6  \\
    \midrule
    Grounding DINO 1.5 Edge \cite{GroundingDINO1.5} & EfficientViT-L1 & Grounding-20M & 800 × 1333 & 45.0    & 36.2  & 33.2  & 36.6  & 36.3  &       & 29.3  & 28.1  & 27.6  & 31.6 \\
    Grounding DINO 1.5 Edge* & EfficientViT-L1 & O365,GoldG,V3Det ($\rm \approx1.56 M$) & 800 × 1333 & 44.6  & 33.1  & 35.9  & 36.8  & 29.4  &       & 27.2  & 32.4  & 27.3  & 24.8  \\
    Dynamic-DINO (Ours) & EfficientViT-L1 & O365,GoldG,V3Det ($\rm \approx1.56 M$) & 800 × 1333 & 46.2  & 36.2  & 41.9  & 39.9  & 31.9  &       &  29.6  &  35.4 &  29.2  &  27.3  \\
    \bottomrule
    \end{tabular}%
    }
  \label{tab:main}%
  \vspace{-0.3cm}
\end{table*}%

\vspace{1mm}
\noindent \textbf{Loss Function.}
The total loss $\mathcal{L}_{total}$ comprises the detection loss $\mathcal{L}_{det}$ and the auxiliary loss $\mathcal{L}_{aux}$, expressed as:
\begin{equation}
    \mathcal{L}_{total}=\mathcal{L}_{det}+\alpha \cdot \mathcal{L}_{aux}
    \label{total_loss}
\end{equation}
where $\alpha$ is balancing coefficient of $\mathcal{L}_{aux}$.
$\mathcal{L}_{det}$ consists of bounding box regression and classification losses. Following the DETR-like work \cite{zhang2022dino}, the L1 loss and GIOU loss are used for bounding box regression branch. For the classification branch, we utilize focal loss as a contrastive loss between the predicted boxes and language tokens. Thus, $\mathcal{L}_{det}$ is calculated as:
\begin{equation}
    \mathcal{L}_{det}=\mathcal{L}_{1}+\mathcal{L}_{\mathrm{GIOU}}+\mathcal{L}_{\mathrm{Focal}}
    \label{det_loss}
\end{equation}
During MoE-Tuning, it is necessary to employ load balancing loss to ensure that each expert is fully utilized. Following MoE-LLaVA \cite{lin2024moe}, we incorporate the load balancing loss into each sparse MoE layer in our Cross-Modality MoE Decoder, which is formulated as:
\begin{equation}
    \mathcal{L}_{aux}=kN \cdot \sum_{i=1}^{kN} \mathcal{F}_i \cdot \mathcal{P}_i
    \label{aux_loss}
\end{equation}
where $kN$ is number of experts, $\mathcal{F}_i$ represents the fraction of tokens processed by each expert $\mathrm{E}_i$, and $\mathcal{P}_i$ represents the average routing probabilities assigned to expert $\mathrm{E}_i$.


\section{Experiments}
\label{sec:Experiments}
\subsection{Experimental Setup}
\label{exp_set}
\noindent \textbf{Pre-training Data.} 
Our Dynamic-DINO is trained on detection and grounding datasets including Objects365 (V1) \cite{O365}, GoldG \cite{GoldG} and V3Det \cite{V3Det} datasets. 
Following \cite{GLIP}, we exclude the images from the COCO dataset in GoldG (GQA \cite{GQA} and Flickr30k \cite{Flickr30k}).

\noindent \textbf{Benchmark.} 
We evaluate the performance of the proposed Dynamic-DINO under a zero-shot setting on the COCO \cite{coco}, LVIS \cite{LVIS} and ODinW \cite{GLIP}. 
Following previous methods \cite{GLIP, GroundingDINO1.0}, we use the standard Average Precision (AP) to evaluate the performance of COCO and ODinW, and the Fixed AP \cite{dave2021evaluating} on LVIS for fair comparison. 

\noindent \textbf{Implementation Details.} 
Dynamic-DINO builds upon the reproduced Grounding DINO 1.5 Edge. We leveraged EfficientViT-L1 as the image backbone, and BERT-base from Hugging Face as the text backbone. We extract three image feature scales, from 8$\times$ to 32$\times$, and downsample the 32$\times$ feature map to 64$\times$ as an extra feature scale. 
By default, we set the number of queries to 900, with 6 decoder layers.
For pre-training stage, we adopt the AdamW, with a base learning rate of 4e-5 for all model parameters expect the image backbone and text backbone, which has a learning rate of 4e-6. The total batch size is 128. The weights allocated to $\mathcal{L}_{\mathrm{Focal}}$, $\mathcal{L}_{1}$ and $\mathcal{L}_{\mathrm{GIOU}}$ are 2.0, 5.0 and 2.0, respectively. Pre-training stage are conducted for 7 epochs.
For MoE-Tuning stage, we initialize the parameters from the pre-trained base model. 
MoE-Tuning stage are conducted for 10 epochs. 
The balancing coefficient $\alpha=0.01$.
All the models are trained on 8 NVIDIA 3090 GPUs.

\begin{table}[!t]
  \centering
  \caption{\textbf{Comparison of zero-shot performance on ODinW.} Dynamic-DINO×16-Top2 model is utilized.}
  \resizebox{0.495 \textwidth}{!}{
    \begin{tabular}{llcc}
    \toprule
    \multicolumn{1}{c}{Model} & \multicolumn{1}{c}{Pre-training Data} & ODinW13 & ODinW35 \\
    \midrule
    Grounding DINO 1.5 Edge* & O365,GoldG,V3Det & 45.8 & 19.6 \\
    Dynamic-DINO (Ours) & O365,GoldG,V3Det & 46.8 & 20.0 \\
    \bottomrule
    \end{tabular}%
    }
  \label{tab:odinw}%
  \vspace{-0.3cm}
\end{table}%

\begin{table}[!t]
  \centering
  \caption{\textbf{Comparison of inference speed.} Dynamic-DINO×16-Top2 model is utilized. FPS is tested on a single A100 40G GPU.}
  \resizebox{0.495 \textwidth}{!}{
    \begin{tabular}{llcc}
    \toprule
    Method & Test Size & \multicolumn{1}{c}{FPS-Pytorch} & \multicolumn{1}{c}{FPS-TensorRT FP32} \\
    \midrule
    Grounding DINO 1.5 Edge & 640 × 640 &  21.7  &  111.6 \\
    Grounding DINO 1.5 Edge* & 640 × 640 & 20.2 &  108.9 \\
    Dynamic-DINO & 640 × 640 & 17.1 & 98.0 \\
    \midrule
    Grounding DINO 1.5 Edge & 800 × 1333 &  18.5  & 75.2 \\
    Grounding DINO 1.5 Edge* & 800 × 1333 & 18.1 & 74.9 \\
    Dynamic-DINO & 800 × 1333 & 15.1 & 66.9 \\
    \bottomrule
    \end{tabular}%
    }
  \label{tab:speed}%
  \vspace{-0.3cm}
\end{table}%

\begin{figure*}[!t] 
    \centering
    \includegraphics[width=0.9\textwidth]{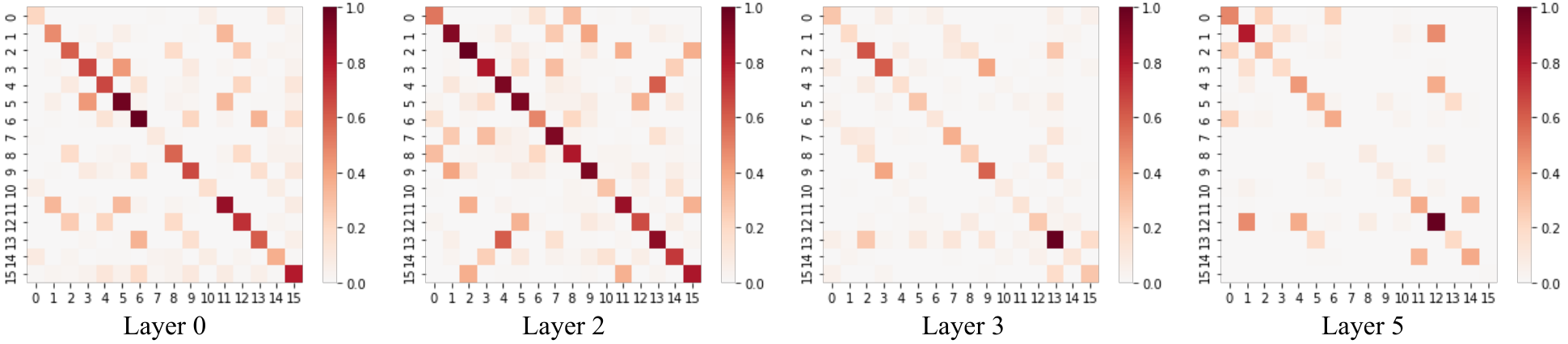}
    \vspace{-0.2cm}
    \caption{\textbf{Expert collaboration.} The normalized co-selection frequencies are quantified for all expert pairs on LVIS-minival \cite{LVIS} with Dynamic-DINO×16-Top2 model, which comprises 16 experts and activates 2 experts per inference.}
    \label{expert_collaboration}
    \vspace{-0.2cm}
\end{figure*}

\begin{figure*}[!t] 
    \centering
    \includegraphics[width=0.9\textwidth]{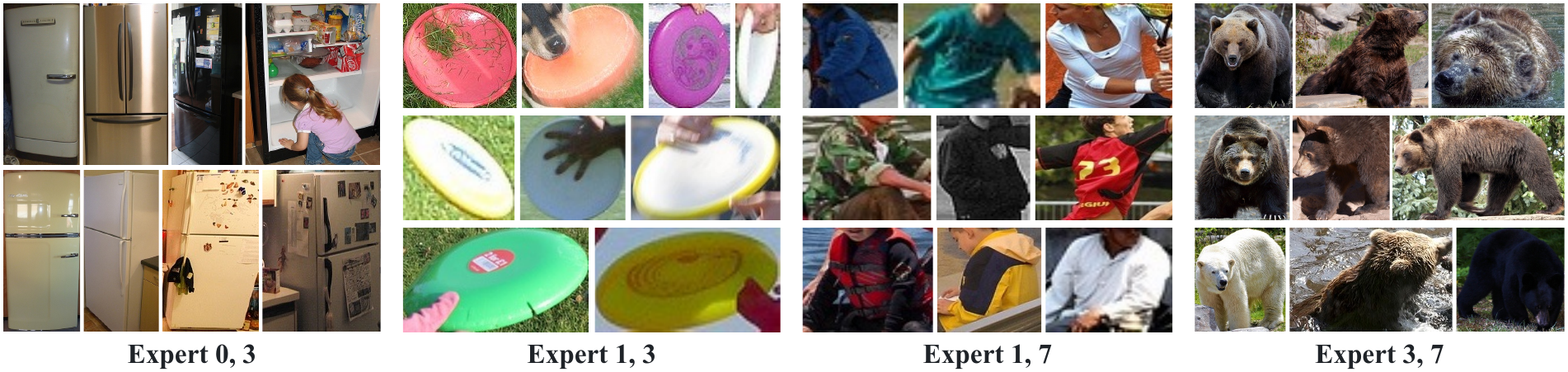}
    \vspace{-0.2cm}
    \caption{\textbf{Token routing examples for COCO.} Image examples of how patches are routed at the MoE layer in the last block of the decoder for the Dynamic-DINO×16-Top2 model. Distinct expert combinations are specialized in processing specific patterns.}
    \label{expert_img}
    \vspace{-0.2cm}
\end{figure*}

\begin{figure}[!t] 
    \centering
    \includegraphics[width=0.45\textwidth]{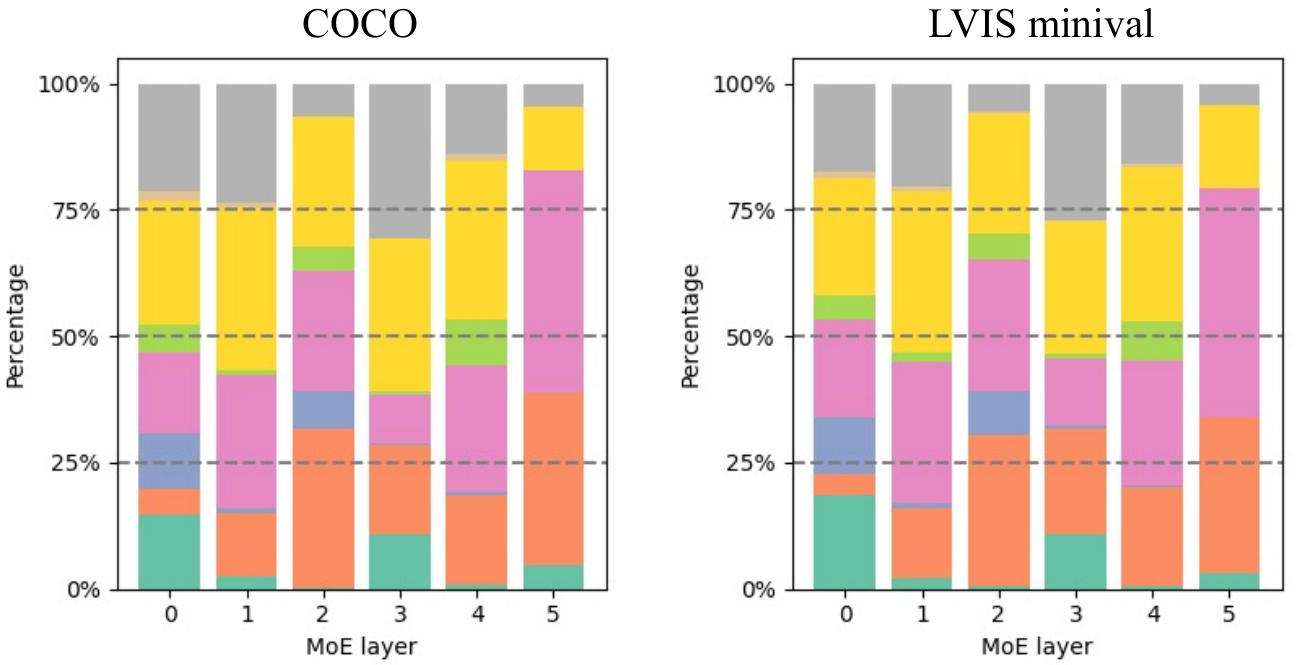}
    \vspace{-0.2cm}
    \caption{\textbf{Distribution of expert loadings.} The workload among experts is quantified with Dynamic-DINO×8-Top2 model during inference on COCO-val and LVIS-minival benchmarks, where each color represents one expert.}
    \label{expert_loadings}
    \vspace{-0.4cm}
\end{figure}

\begin{table*}[htbp]
  \centering
  \caption{\textbf{Ablation study of tuning the parameters of different subsets.} Dynamic-DINO×16-Top2 model is utilized. Image resolution is 640 × 640. Feature Enhancer specifically denotes the cross-attention module within it. We examine the performance of fine-tuning different parts of the parameters while keeping other modules frozen.}
  \resizebox{0.8 \textwidth}{!}{
    \begin{tabular}{ccc|cccccccccc}
    \toprule
    \multicolumn{1}{l}{\multirow{2}[2]{*}{MoE Layer}} & \multicolumn{1}{l}{\multirow{2}[2]{*}{Feature Enhancer}} & \multicolumn{1}{l|}{\multirow{2}[2]{*}{Detection Head}} & COCO-val & \multicolumn{4}{c}{LVIS-minival} &       & \multicolumn{4}{c}{LVIS-val} \\
\cmidrule{5-8}\cmidrule{10-13}          &       &       & $\rm AP_{box}$ &  $\rm AP_{all}$ &$\rm AP_{r}$ & $\rm AP_{c}$ & $\rm AP_{f}$ &       & $\rm AP_{all}$ &$\rm AP_{r}$ & $\rm AP_{c}$ & $\rm AP_{f}$ \\
    \midrule
    \checkmark     &       &       & 43.4  & 32.4  &  37.3  &  35.6  & 28.7 &       & 26.2  &  31.8  &   25.9  & 24.1  \\
    \checkmark     & \checkmark     &       & 43.5  & 32.7  &  35.6  & 36.1  &  29.2  &       & 26.7  &  31.7  &   26.6  &  24.7 \\
    \checkmark     &       & \checkmark     & 43.4  & 33.4  &  37.8  &   36.3 &  30.0  &       & 27.5  &    33.7 &   27.1  & 25.4 \\
    \checkmark     & \checkmark     & \checkmark     & 43.7  & 33.6  &   37.0  &  36.6  &     30.3  &       & 27.4  &  32.4  &  26.9  & 25.6  \\
    \bottomrule
    \end{tabular}%
    }
  \label{tab:tuning_pos}%
  \vspace{-0.3cm}
\end{table*}%

\subsection{Comparisons with the State-of-the-art}
\label{compare}
For a comprehensive evaluation, we compare our Dynamic-DINO with the state-of-the-art real-time open-vocabulary detectors, including YOLO-World v2 \cite{Yolo-world}, OmDet-Turbo \cite{zhao2024real}, OVLW-DETR \cite{wang2024ovlw} and Grounding DINO 1.5 Edge \cite{GroundingDINO1.5}.
As reported in Tab. \ref{tab:main}, Dynamic-DINO achieves comparable performance with the official Grounding DINO 1.5 Edge across different resolutions. Notably, Dynamic-DINO significantly enhances the detection performance on rare classes, indicating that MoE-Tuning effectively alleviates the long-tail problem.
Since the official Grounding DINO 1.5 Edge did not report performance on ODinW, we only compared the performance of our reproduced Grounding DINO 1.5 Edge and Dynamic-DINO in Tab. \ref{tab:odinw}.
Additionally, the speed comparison is reported in Tab. \ref{tab:speed}. Due to its closed-source status, the reproduced Grounding DINO 1.5 Edge is slightly slower than the official version. After MoE-Tuning, there is a minor decrease in inference speed because the current implementation feeds tokens forward to different expert networks in a sequential loop, significantly reducing efficiency. Future work will optimize this engineering problem for acceleration.

\subsection{Statistical Analysis}
\noindent \textbf{Routing Distributions.} In Fig. \ref{expert_loadings}, we present the statistical results about the expert loading during inference through Dynamic-DINO×8-Top2 on COCO-val and LVIS-minival benchmarks, where each color represents one expert.
The dynamic selection of experts varies notably across different layers, indicating that experts have learned a certain mechanism to divide the task in a specific manner.

\noindent \textbf{Expert Collaboration.} Fig. \ref{expert_collaboration} provides further insights into the collaborative dynamics among the experts through Dynamic-DINO×16-Top2.
We quantify the co-selection frequency for all possible expert pairs on the LVIS-minival benchmark and applied normalization for the results.
In the shallow layers, experts tend to cooperate with a diverse range of peers to explore a wider search space. In contrast, in the deeper layers, experts gradually refine their preferences, focusing on consistent collaborations with 2-3 specific partners to process distinct patterns.

\noindent \textbf{Token Routing Examples.} Fig. \ref{expert_img} provides a visualization of the routing mechanism for image patches at the MoE layer in the last decoder block.
The results reveal that distinct expert combinations are specialized in processing specific patterns. For example, experts 0 and 3 mainly manage tokens related to refrigerators, whereas experts 1 and 7 are dedicated to tokens associated with clothing.
These findings confirm our hypothesis that tokens with similar patterns tend to select identical expert combinations. 
Consequently, a more fine-grained division of experts enables a broader expert combinations, thereby reducing the number of patterns handled by each expert group.
This inherent efficiency explains how we achieved superior network performance with relatively limited data.

\subsection{Ablation Study}

\noindent \textbf{Effect of Tuning the Parameters of Different Subsets.} 
The results in Tab. \ref{tab:tuning_pos} demonstrate that the detection head plays a critical role in the MoE-Tuning process, achieving a significant improvement of +1.3 AP on the LVIS-val.
In addition, jointly fine-tuning the cross-attention in feature enhancer enables further performance gains.

\begin{figure}[!t] 
    \centering
    \includegraphics[width=0.4\textwidth]{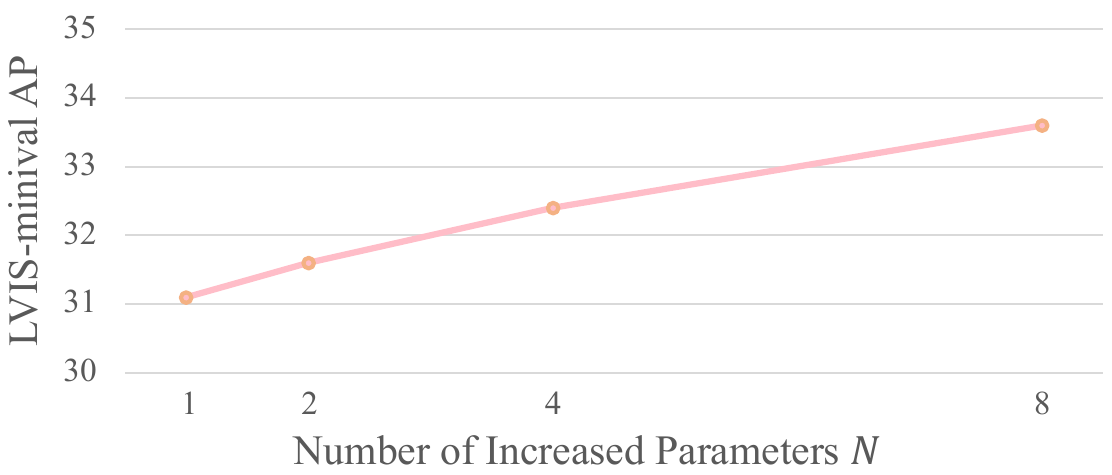}
    \caption{\textbf{Effect of parameter quantity.} The horizontal axis $N$ represents scaling the FFN to $N$ units.}
    \label{parameter_quantity}
\end{figure}

\begin{figure}[!t] 
    \centering
    \includegraphics[width=0.4\textwidth]{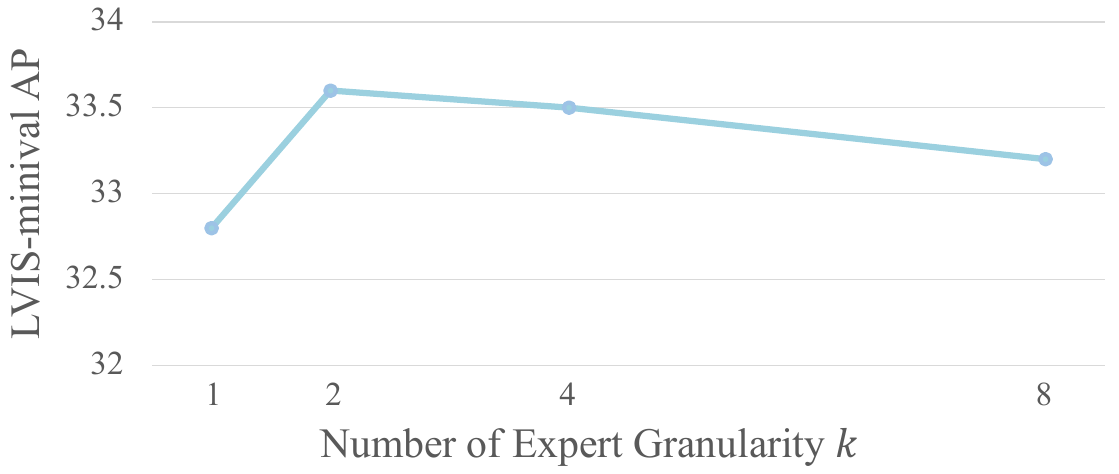}
    \caption{\textbf{Effect of expert granularity.} The horizontal axis $k$ denotes decoupling a FFN into $k$ partitions and $N=8$ is utilized.}
    \label{expert_granularity}
\end{figure}

\noindent \textbf{Effect of the Search Space.} Fig. \ref{parameter_quantity} suggests that a larger parameter quantity consistently yields performance improvements.
Meanwhile, with fixed parameters, decoupling a single FFN into two experts further enhances performance, but excessive subdivision causes a decline, as shown in Fig. \ref{expert_granularity}.
We attribute this to the limited training data, where an excessively large search space increases overfitting risk, compromising zero-shot performance.

\noindent \textbf{Effect of the Training Efficiency.} 
As shown in Tab. \ref{tab:training_efficiency}, under the same training data and GPU conditions, MoE-Tuning achieves a 1.87$\times$ speedup compared with the pre-training scheme.
In addition, extended pre-training offers marginal performance improvements, while MoE-Tuning enables substantial enhancements, illustrated in Fig. \ref{training_coco}.



\noindent \textbf{Effect of the Datasets.} While Dynamic-DINO delivers strong results with limited data, Tab. \ref{tab:data} reveals that its performance grows markedly with increased training data. It is worth noting that all datasets used in this work are open-source, ensuring reproducibility and accessibility.






\section{Limitation Discussion}
\label{sec:Limitations}
This work builds upon the Grounding DINO 1.5 Edge as the base model, extending it from a dense model to a dynamic inference model based on MoE-Tuning. With limited open-source data, our method matches the performance of official Grounding DINO 1.5 Edge. However, due to computational constraints, limited to 8 NVIDIA 3090 GPUs, we are unable to train and validate our method on the scaled-up Grounding DINO 1.5 Pro model, nor explore the performance boundaries of MoE-Tuning with sufficient datasets. 
Parallel acceleration of the multi-expert feed-forward process also requires further refinement in the future.

\begin{table}[!t]
  \centering
  \caption{\textbf{Comparison of training efficiency.} Dynamic-DINO×16-Top2 model is utilized. Image resolution is 640 × 640.}
  \resizebox{0.49 \textwidth}{!}{
    \begin{tabular}{llcc}
    \toprule
    Method & Pre-training Data & GPUs & \multicolumn{1}{c}{Training Time / Epoch} \\
    \midrule
    Pre-Training & O365,GoldG,V3Det &  8 RTX-3090 & 14.0h \\
    MoE-Tuning  & O365,GoldG,V3Det &  8 RTX-3090 & 7.5h \\
    \bottomrule
    \end{tabular}%
  }
  \label{tab:training_efficiency}%
\end{table}%

\begin{table}[!t]
  \centering
  \caption{\textbf{Ablation study of training datasets.} Dynamic-DINO×16-Top2 model is utilized. Image resolution is 640 × 640.}
  \resizebox{0.495 \textwidth}{!}{
    \begin{tabular}{c|cccccccccc}
    \toprule
    \multirow{2}[4]{*}{Training Data}  & COCO-val & \multicolumn{4}{c}{LVIS-minival} &       & \multicolumn{4}{c}{LVIS-val} \\
\cmidrule{3-6}\cmidrule{8-11}      & $\rm AP_{box}$ &  $\rm AP_{all}$ &$\rm AP_{r}$ & $\rm AP_{c}$ & $\rm AP_{f}$ &     &  $\rm AP_{all}$ &$\rm AP_{r}$ & $\rm AP_{c}$ & $\rm AP_{f}$ \\
    \midrule
    O365  & 45.8  & 21.4  &  26.6  &  22.2  & 19.7 &     & 16.6  &  20.5  & 14.7  & 16.9  \\
    O365,GoldG  & 45.8  & 33.9  &  42.3  & 36.5  &  30.0  &     & 27.1  &  32.4 & 26.3 &  25.5 \\
    O365,GoldG,V3Det  & 46.2  & 36.2  &  41.9  & 39.9 &  31.9  &    & 29.6  &  35.4 & 29.2  & 27.3 \\
    \bottomrule
    \end{tabular}%
    }
  \label{tab:data}%
\end{table}%

\section{Conclusion}
\label{sec:Conclusion}
In this paper, we propose Dynamic-DINO, a novel framework that explores the integration of real-time open-vocabulary object detection with Mixture of Experts (MoE). We demonstrate that diverse expert combinations can adaptively process specific patterns. Thus, Dynamic-DINO only activates the relevant experts based on the input data patterns during inference, achieving impressive performance even with limited training data. 
Specifically, Dynamic-DINO builds upon our reproduced Grounding DINO 1.5 Edge, extending it from a dense model into a dynamic inference framework via MoE-Tuning.
Additionally, we design a granularity decomposition mechanism to segment expert networks, expanding the subnet search space while strictly maintaining the activated parameters equivalent to those of a single FFN in the base model. To prevent performance degradation at the start of fine-tuning, we further propose a pre-trained weight allocation strategy for the experts, coupled with specific router initialization. Extensive experiments validate the effectiveness of our proposed method.

\section{Acknowledgement}
\label{sec:Acknowledgement}
This work is supported in part by National Science Foundation for Distinguished Young Scholars under Grant 62225605, Project 12326608 supported by NSFC, Zhejiang Provincial Natural Science Foundation of China under Grant LD24F020016, Ningbo Science and Technology Special Projects under Grant No. 2025Z028, and the Fundamental Research Funds for the Central Universities.
{
    \small
    \bibliographystyle{ieeenat_fullname}
    \bibliography{main}
}
\clearpage
\setcounter{page}{1}
\maketitlesupplementary
\setcounter{section}{0}
\renewcommand\thesection{\Alph{section}}

\section{Appendix}
\label{sec:Appendix}
\subsection{Datasets Details}
Tab. \ref{tab:pre-Training data} presents the dataset specifications utilized for pre-training Dynamic-DINO, including the Objects365 (V1) \cite{O365}, GQA \cite{GQA}, Flickr30k \cite{Flickr30k}, and V3Det \cite{V3Det} datasets, where Texts denotes the number of categories for the detection dataset and the number of phrases for the grounding dataset, Images denotes the number of images and Annotation denotes the number of instance annotations. The total number of samples in our pre-training dataset is 1.56M.

\begin{table}[htbp]
  \caption{\textbf{Pre-Training Data.}}
  \resizebox{0.48 \textwidth}{!}{
    \begin{tabular}{llccc}
    \toprule
    Dataset & Type & Texts & Images & Annotation \\
    \midrule
    O365 \cite{O365} & Detection & 365 & 609K & 9621K \\
    V3Det \cite{V3Det} & Detection & 13K & 184K & 1233K \\
    GQA \cite{GQA} & Grounding & 387K & 621K & 3681K \\
    Flickr30k \cite{Flickr30k} & Grounding & 94K & 149K & 641K \\
    \bottomrule
    \end{tabular}%
  }
  \label{tab:pre-Training data}%
\end{table}%

\subsection{Core Codes}
The core implementation of our MoE-Tuning is detailed in Algorithm \ref{code}, encompassing expert initialization and router initialization.
Following MoE \cite{fedus2022switch} paradigm, we scale up the model by expanding the FFN in each layer of the decoder into $N$ FFNs of identical size.
For each FFN, its intermediate hidden dimension is evenly divided into $k$ partitions, thereby constructing $k\times N$ experts.
In addition, we initialize the experts by assigning the pre-trained FFN weights from the base model to each expert.
For router initialization, we first randomly initialize the weights $W'_r \in \mathbb{R}^{N\times D}$, and then replicate each centroid vector in $W'_r$ $k$ times to form the router weights $W_r \in \mathbb{R}^{kN\times D}$.
With this initialization, the router is guaranteed to select the $k$ experts derived from the same FFN at the start of fine-tuning, ensuring incremental performance improvements during MoE-Tuning.

\begin{algorithm}[htbp]
\caption{MoE Initialization}
\label{code}
\begin{lstlisting}[language=Python]
"""
Input:
n: int
k: int
ffn: nn.Module
"""
embed_dim = ffn.embed_dim
ffd_dim = ffn.ffd_dim // k

ffns = [
    FFN(embed_dim, ffd_dim) 
    for _ in range(k)
]
for i in range(k):
    ffns[i].w1 
      =ffn.w1[i*ffd_dim:(i+1)*ffd_dim,:]
    ffns[i].b1 
      =ffn.b1[i*ffd_dim:(i+1)*ffd_dim]
    ffns[i].w2 
      =ffn.w2[:,i*ffd_dim:(i+1)*ffd_dim]
    ffns[i].b2 = ffn.b2 / k
    
self.experts = nn.ModuleList([])
for i in range(n):
    for j in range(k):
        self.experts.append(
            copy.deepcopy(ffns[j])
        )

w_gate = torch.randn(n, 1, embed_dim)
w_gate = w_gate.repeat(1, k, 1)
w_gate = w_gate.reshape(n*k, embed_dim)
self.router = nn.Parameter(
    w_gate, requires_grad=True)
\end{lstlisting}
\end{algorithm}

\subsection{More Experiments}
\noindent \textbf{Ablation Study on Parameter Numbers.} Our method can flexibly adjust total parameters while keeping activated parameters unchanged. As shown in Table \ref{tab:parameter}, even +6M parameters bring +0.73 AP on average, with scaling parameters yielding greater improvements.

\begin{table}[htbp]
  \centering
  \caption{Comparison of the parameter numbers. All models are trained on O365, GoldG, and V3Det. Image resolution is 640 $\times$ 640. ``Parameters" represents active parameters / total parameters. Dynamic-DINO×N-Top2 indicates a model with N experts, where 2 experts are activated per inference.}
  \resizebox{0.495 \textwidth}{!}{
    \begin{tabular}{lcccc}
    \toprule
    Method & Parameters & COCO-val & LVIS-minival & LVIS-val \\
    \midrule
    G-DINO 1.5 Edge & 178M/178M & 42.6  & 31.1  & 25.4  \\
    Dynamic-DINO×4-Top2 & 178M/184M & 43.2(\textcolor{red}{+0.6})  & 31.6(\textcolor{red}{+0.5})  & 26.5(\textcolor{red}{+1.1})  \\
    Dynamic-DINO×8-Top2 & 178M/197M & 43.4(\textcolor{red}{+0.8})  & 32.4(\textcolor{red}{+1.3})  & 26.9(\textcolor{red}{+1.5})  \\
    Dynamic-DINO×16-Top2 & 178M/222M & 43.7(\textcolor{red}{+1.1})  & 33.6(\textcolor{red}{+2.5})  & 27.4(\textcolor{red}{+2.0})  \\
    \bottomrule
    \end{tabular}%
  }
  \label{tab:parameter}%
\end{table}%

\noindent \textbf{Ablation Study on MoE Deployment.} As shown in Table \ref{tab:image_encoder}, extending MoE layers to FFN in image encoder, the performance further increases by +0.5 AP on average.

\begin{table}[htbp]
  \centering
  \tiny
  \caption{Ablation study of MoE deployment across model parts. Dynamic-DINO×16-Top2 is utilized. All models are trained on O365, GoldG, and V3Det. Image resolution is 800 $\times$ 1333.}
  \renewcommand{\arraystretch}{1.0}
  \resizebox{0.495 \textwidth}{!}{
    \begin{tabular}{cc|ccc}
    \toprule
    Decoder & Image Encoder & COCO-val & LVIS-minival & LVIS-val \\
    \midrule
    - & - & 42.6  & 31.1  & 25.4  \\
    \checkmark & - & 43.7(\textcolor{red}{+1.1})  & 33.6(\textcolor{red}{+2.5})  & 27.4(\textcolor{red}{+2.0})  \\
    \checkmark & \checkmark & \textbf{44.5}(\textcolor{red}{+1.9})  & \textbf{33.7}(\textcolor{red}{+2.6})  & \textbf{28.0}(\textcolor{red}{+2.6})  \\
    \bottomrule
    \end{tabular}%
  }
  \label{tab:image_encoder}%
\end{table}%

\noindent \textbf{Ablation Study on Model Initialization.} We validate the effectiveness of our initialization modification. As shown in Table \ref{tab:init}, it boosts the accuracy ceiling.

\begin{table}[htbp]
  \centering
  \caption{Ablation study for the initialization. Dynamic-DINO×16-Top2 is utilized. All models are trained on O365, GoldG, and V3Det. Image resolution is 640 $\times$ 640.}
  \resizebox{0.495 \textwidth}{!}{
    \begin{tabular}{lccc}
    \toprule
    Method & COCO-val & LVIS-minival & LVIS-val \\
    \midrule
    G-DINO 1.5 Edge & 42.6  & 31.1  & 25.4  \\
    Dynamic-DINO w/o Initialization & 43.1  & 32.5  & 26.2  \\
    \rowcolor{gray!15} Dynamic-DINO w/ Initialization & \textbf{43.7}  & \textbf{33.6}  & \textbf{27.4}  \\
    \bottomrule
    \end{tabular}%
  }
  \label{tab:init}%
\end{table}%

\noindent \textbf{Results on RefCOCO.} Experiments on RefCOCO, RefCOCO+ and RefCOCOg are added in Table \ref{tab:RefCOCO}. Results show that our method still works on zero-shot REC tasks.

\begin{table}[htbp]
  \centering
  \caption{Comparison of zero-shot performance on RefCOCO, RefCOCO+ and RefCOCOg. All models are trained on O365, GoldG, and V3Det. Image resolution is 640 $\times$ 640.}
  \resizebox{0.495 \textwidth}{!}{
    \begin{tabular}{l|ccc|ccc|cc}
    \toprule
    \multicolumn{1}{c|}{\multirow{2}[2]{*}{Method}} & \multicolumn{3}{c|}{RefCOCO} & \multicolumn{3}{c|}{RefCOCO+} & \multicolumn{2}{c}{RefCOCOg} \\
          & val   & testA & testB & val   & testA & testB & val   & test \\
    \midrule
    G-DINO 1.5 Edge & 43.8  & 49.9  & 39.5  & 43.3  & 47.9  & 40.2  & 51.2  & 52.8  \\
    \rowcolor{gray!15} Dynamic-DINO (Ours) & \textbf{47.9}  & \textbf{53.9}  & \textbf{42.2}  & \textbf{47.4}  & \textbf{52.0}  & \textbf{42.3}  & \textbf{56.6}  & \textbf{56.5}  \\
    \bottomrule
    \end{tabular}%
  }
  \label{tab:RefCOCO}%
\end{table}%

\noindent \textbf{Performance Comparisons on Edge Devices.} We evaluate the pre-trained model on Jetson Orin NX SUPER 8GB. As shown in Table \ref{tab:edge}, our method introduces only +0.24M FLOPs and -0.8 FPS over the baseline while achieving +1.87 AP on average.

\begin{table}[htbp]
  \centering
  \caption{Performance comparisons on NVIDIA Orin NX. All models are trained on O365, GoldG, and V3Det. Image resolution is 640 $\times$ 640. Dynamic-DINO×16-Top2 is utilized. FLOPs are measured solely for the Decoder, which contains the MoE Layers in our method. FPS evaluates the full feed-forward pass.}
  \resizebox{0.495 \textwidth}{!}{
    \begin{tabular}{lccccc}
    \toprule
    Method & COCO-val & LVIS-minival & LVIS-val & FLOPs & FPS \\
    \midrule
    G-DINO 1.5 Edge & 42.6  & 31.1  & 25.4  & 2679.51M & 10.2  \\
    \rowcolor{gray!15} Dynamic-DINO (Ours) & 43.7  & 33.6  & 27.4  & 2679.75M & 9.4  \\
    \bottomrule
    \end{tabular}%
  }
  \label{tab:edge}%
\end{table}%

\subsection{Visualizations}
\begin{figure*}[!t]
    \centering
    \includegraphics[width=0.99\textwidth]{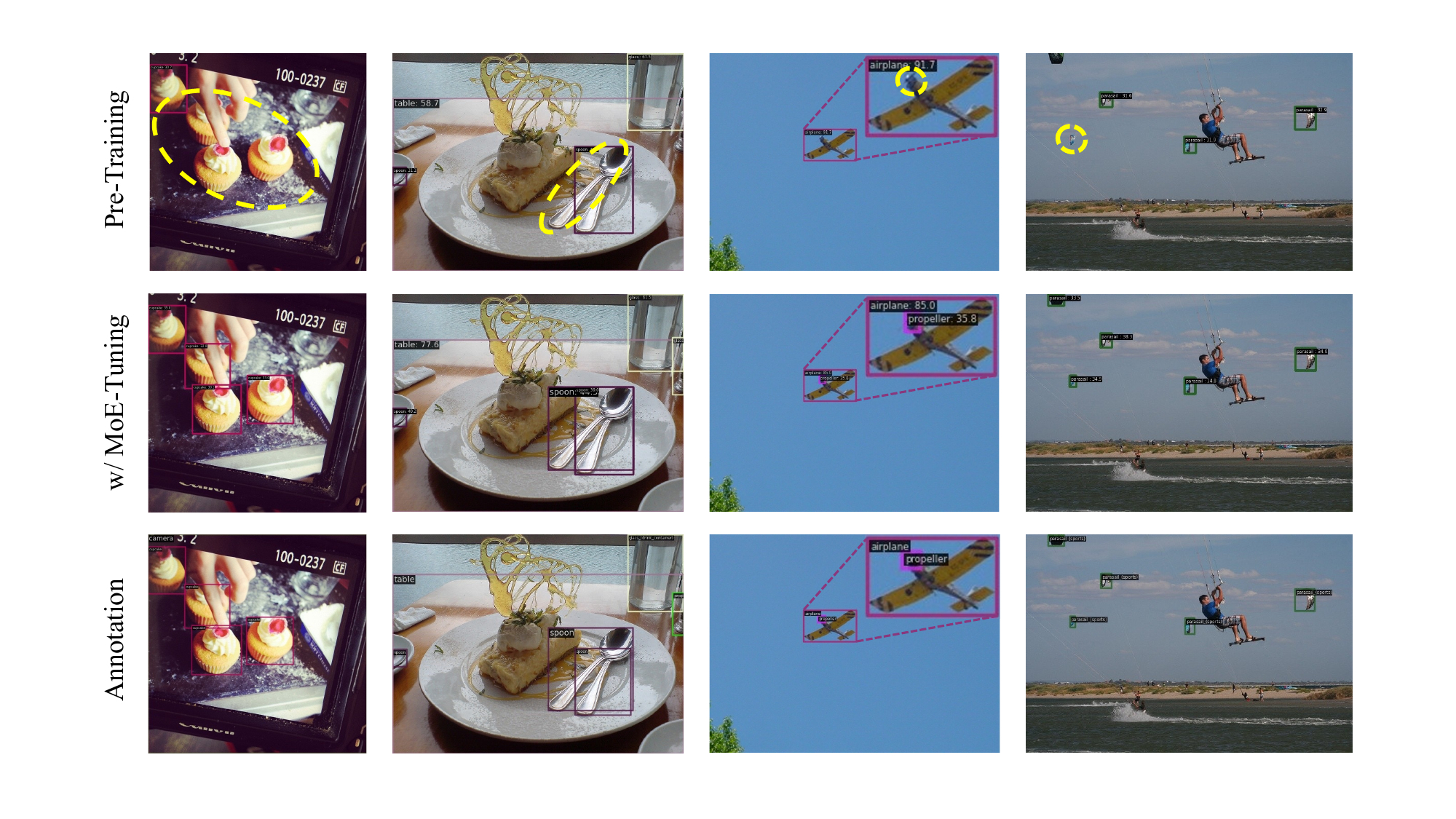}
    \caption{\textbf{Comparison of visualization results for zero-shot inference on LVIS.} We visualize the predictions of our pre-trained base model and Dynamic-DINO after MoE-Tuning. The failures are highlighted with a \textcolor{yellow}{yellow} circle.}
    \label{fig:vis}
\end{figure*}

\begin{figure*}[!t]
    \centering
    \includegraphics[width=0.99\textwidth]{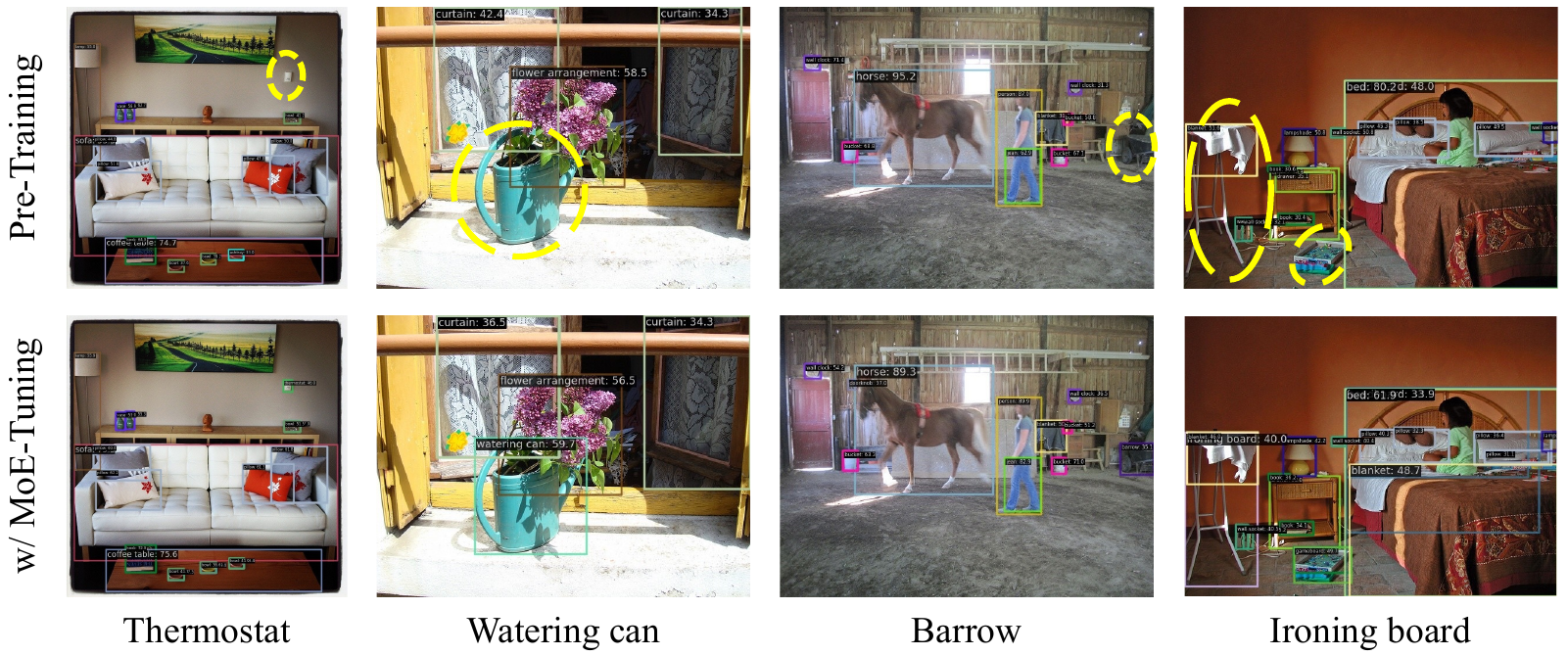}
    \caption{\textbf{Comparison of visualization results for zero-shot inference on rare classes of LVIS.} We visualize the predictions of our pre-trained base model and Dynamic-DINO after MoE-Tuning. The failures are highlighted with a \textcolor{yellow}{yellow} circle.}
    \label{fig:rare_class}
\end{figure*}

Fig. \ref{fig:vis} provides a comparative visualization of the model's zero-shot object detection performance before and after the implementation of MoE-Tuning. The results demonstrate a significant improvement in the model's sensitivity to both object quantity and small-scale targets. 
Fig. \ref{fig:rare_class} further visualizes the improvement in the model's ability to detect rare classes, indicating that MoE-Tuning effectively alleviates the long-tail problem.

\subsection{More Statistical Analysis}
\begin{figure*}[!t]
    \centering
    \includegraphics[width=0.99\textwidth]{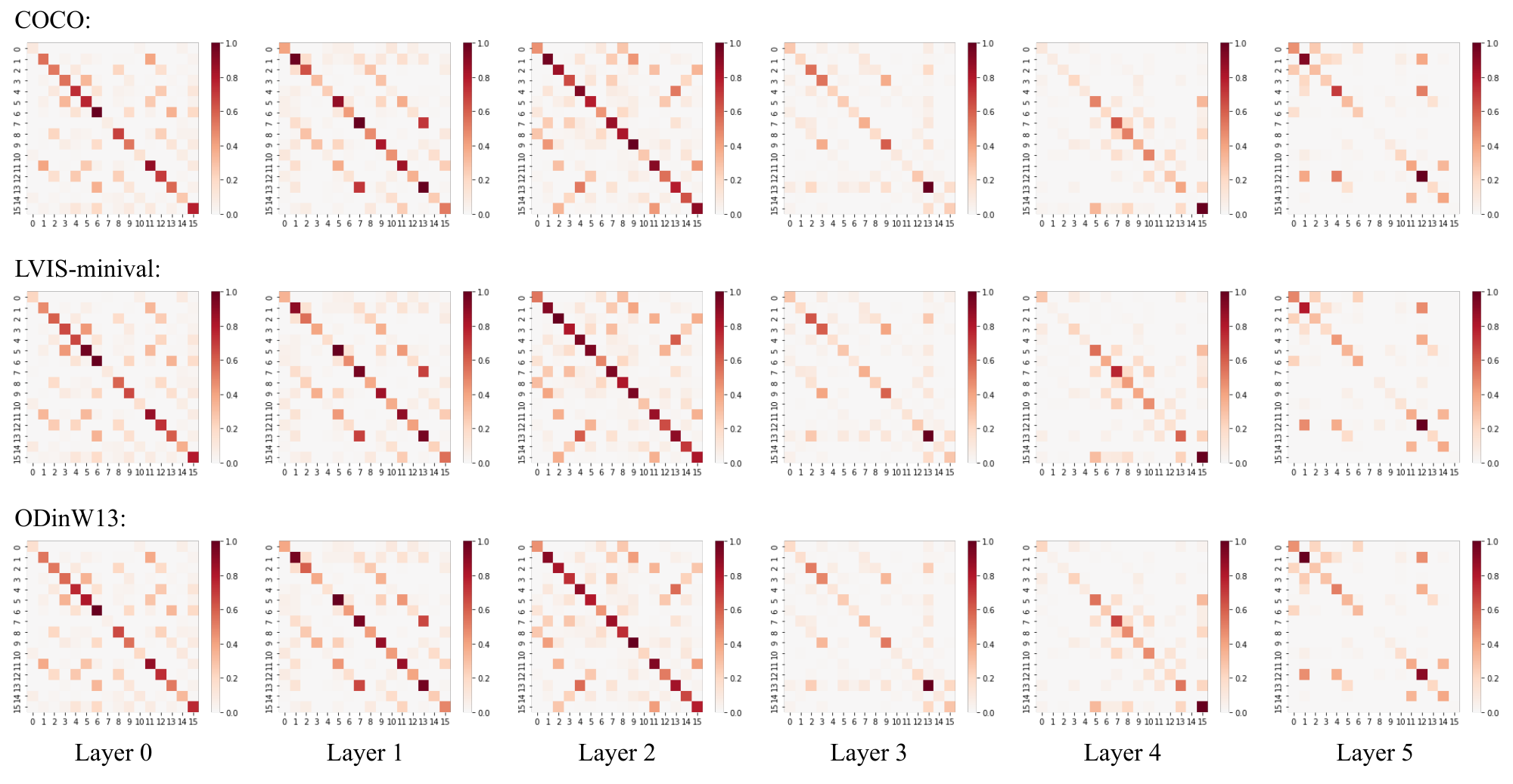}
    \caption{\textbf{Expert collaboration across 3 datasets.} The normalized co-selection frequencies are quantified for all expert pairs with Dynamic-DINO×16-Top2 model, which comprises 16 experts and activates 2 experts per inference.}
    \label{fig:expert_collaboration_all}
\end{figure*}

Fig. \ref{fig:expert_collaboration_all} provides a detailed visualization of the expert collaboration statistics across each MoE layer of Dynamic-DINO, evaluated on the COCO, LVIS-minival, and ODinW13. The results reveal that Dynamic-DINO exhibits a nearly consistent pattern of expert collaboration across diverse datasets, which underscores the stability of expert collaboration and the sufficiency of training.

\end{document}